\documentclass[10pt, a4paper]{article}

\usepackage[final]{lrec2026} %

\usepackage[dvipsnames]{xcolor}
\usepackage{lingex}
\usepackage{times}
\usepackage{latexsym}
\usepackage{booktabs}
\usepackage{caption}

\usepackage{multirow} %
\usepackage{pifont} %
\newcommand{\YES}{\ding{51}} %
\newcommand{\NO}{{\color{gray!50}\ding{55}}} %

\title{Prague Dependency Treebank - Consolidated 2.0: \\ Enriching a Complex Annotation Scheme}

\name{Marie Mikulová, Jiří Mírovský, Milan Straka, Pavlína Synková, Jan Štěpánek,\\ \large\textbf{Barbora Štěpánková, Jan Hajič}}

\address{Charles University, Faculty of Mathematics and Physics, Institute of Formal and Applied Linguistics \\
         Malostranské náměstí 25, 118 00 Prague 1, Czech Republic \\
        \{mikulova,mirovsky,straka,synkova,stepanek,stepenakova,hajic\}@ufal.mff.cuni.cz\\}

\abstract{
The Prague Dependency Treebank framework is unique in its attempt to systematically include and link different layers of language, including a meaning representation with several types of inter-sentential phenomena, especially coreference and discourse relations. We present its second consolidated version (PDT-C 2.0), which concludes almost 30-years long project of sustained development of the resource to a uniformly and coherently annotated, genre-diversified, almost 4 million token language resource of Czech language, with accompanying fully compatible lexicons. In addition to continuous linguistic research, the richly linguistically annotated corpus is also widely used in international comparisons of the development of traditional and novel NLP tools as well as in conversions into other formalisms. The corpus and the trained parsers are available under the CC BY-NC-SA licence.
 \\ \newline  }

\usepackage{fancyhdr}
\fancypagestyle{officialbibref}{\fancyhf{}\fancyheadoffset[L,R]{50pt}\fancyhead[C]{This paper was published at \textbf{LREC 2026} -- please cite the published version {\small\url{https://aclanthology.org/2026.lrec-1.908/}}.}\fancyfoot[C]{\normalsize\thepage}\addtolength{\topmargin}{-50pt}\addtolength{\headsep}{50pt}}
\begin{document}
\thispagestyle{officialbibref}
\pagenumbering{arabic}\pagestyle{plain}

\maketitleabstract

\section{Introduction}

\begin{figure*}[t]
\captionsetup{justification=centering}
\begin{center}
\includegraphics[scale=0.5]{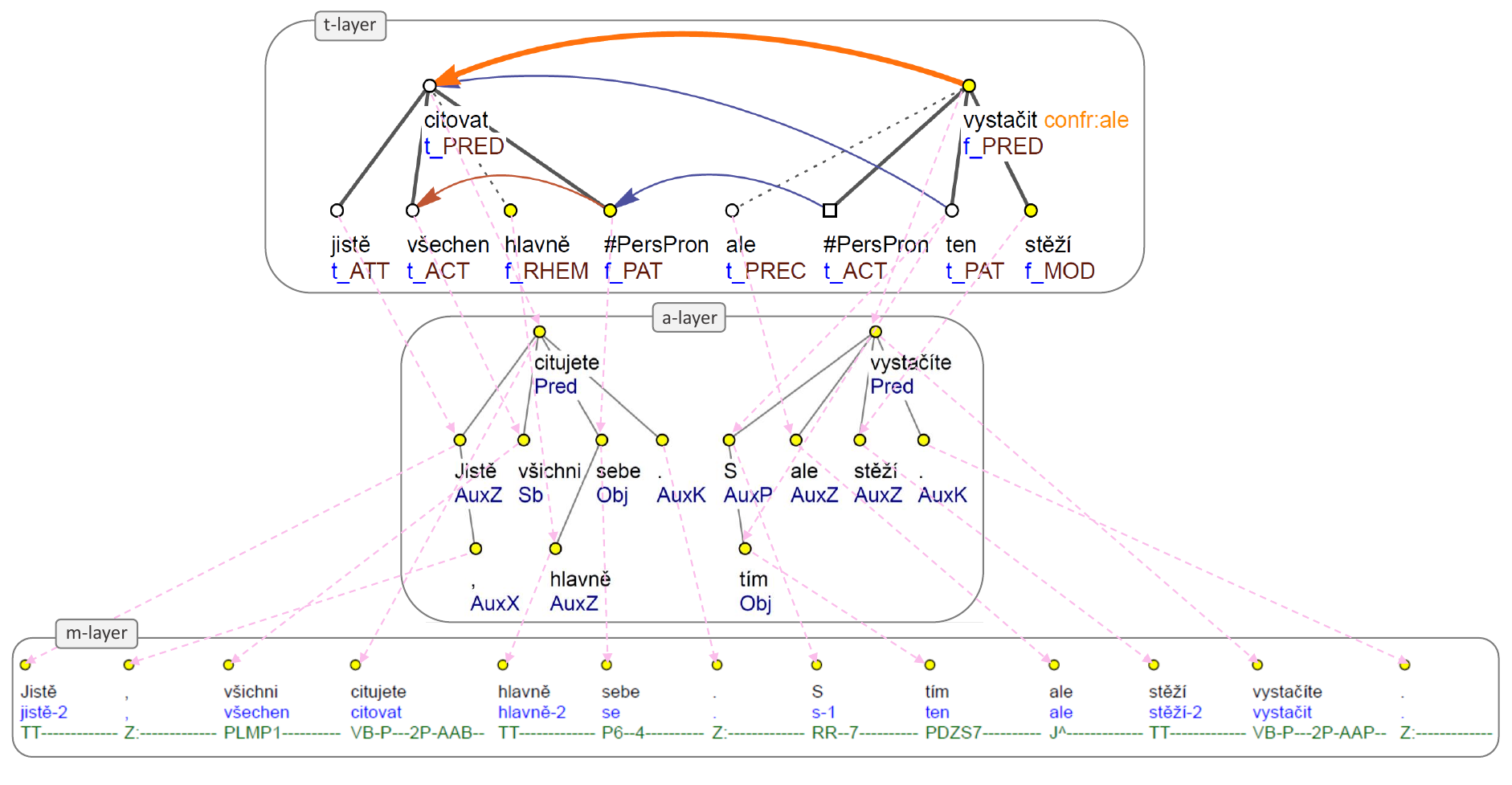}
\caption{Multi-layer annotation scheme in the PDT-C treebank, illustrated on the text:\\
  \textit{Jistě, všichni citujete hlavně sebe. S~tím ale stěží vystačíte.}\\
  Of-course, you-all cite mainly yourself. With that but hardly suffice-you.\\
  ‘Of course, you all mainly cite yourself. But that's hardly enough.’}
\label{fig:layers}
\end{center}
\end{figure*}

We present the \textbf{Prague Dependency Treebank - Consolidated 2.0} (PDT-C 2.0;  \citealplanguageresource{pdtc20})
a second consolidated release of the existing original PDT-corpora of Czech data published in one package to allow for easier data handling. Compared to the previous 1.0 version, the data are now fully manually annotated. Included is a morphological dictionary MorfFlex \citeplanguageresource{morfflex}, and a valency lexicon PDT-Vallex \citeplanguageresource{pdtvallex}. Both external resources are fully compatible with PDT-C annotation. In the paper, we summarize in a balanced manner all the aspects of this large language resource: data genres, multi-layer annotation scheme, various types of rich linguistic annotation, the linked lexical resources and their relation to the annotation. The presented language resource is unique in the following aspects:

\begin{itemize}
\item \textbf{multi-layer architecture}: the complex structure of language is captured through interlinked hierarchical layers of annotation as illustrated in Fig.~\ref{fig:layers}. This enables researchers to process specific linguistic aspects independently, allowing for more detailed and precise analyses. At the same time, the interconnectedness of the layers makes it possible to study how meaning is linked to text. 

\item \textbf{rich linguistic annotation} spanning from morphology and syntax to semantics, including  \textbf{inter-sentential phenomena} (especially coreference, discourse relations). 
See Tab.~\ref{tab:annot} for overview.

\item \textbf{genre-diversified datasets}: written, translated, spoken, and user-generated.
See Tab.~\ref{tab:volume} for the datasets overview.

\item \textbf{large volume of data}: more than 4 million tokens; see Tab.~\ref{tab:volume} for the volume overview.

\item all annotations were performed \textbf{manually}. 
\end{itemize}

PDT-C 2.0 is provided as a digital open resource accessible to all users via LINDAT/CLARIAH-CZ.\footnote{\url{http://hdl.handle.net/11234/1-5813}}

The paper is organized as follows: The PDT multi-layer annotation scheme is described in  Sect.~\ref{layers}; the genre diversity of the data is presented in Sect.~\ref{data}; the voluminous manual annotation is emphasized in Sect.~\ref{volume}; the richness of the linguistic annotation is outlined in Sect.~\ref{rich}; external, fully compatible language resources are mentioned in Sect.~\ref{resources}; the application of the treebank in the field of NLP (parser development and conversion into other frameworks) is presented in Sect.~\ref{related}. We conclude in Sect.~\ref{conclusion}, describing also the future work and ongoing annotation efforts (Sect.~\ref{future}).
 
\subsection{From 1.0 to PDT-C 2.0}
\label{from10to20}

Compared to the previous 1.0 version \cite{pdtc10}, the data are now fully manually annotated. The novelty lies in the following:

\begin{itemize}
    \item  Manual annotation at the \textbf{surface syntax} layer (Sect.~\ref{alayer}) is now performed in those parts of the corpus that were previously annotated only by automatic tools. The goal of the annotation work was also to consolidate the manual annotation across all layers, including previously manually annotated parts. Annotators follow all annotation layers during the annotation process. This resulted in many modifications and corrections to the original annotation.
    \item Manual annotation of \textbf{discourse} relations (Sect.~\ref{discourse}) is now provided for all datasets.
    \item Manual annotation of \textbf{coreference} (Sect.~\ref{coreference}) is now provided for all datasets.
\end{itemize}

\section{Multi-layer Architecture}
\label{layers}

The PDT annotation scheme is based on the well-developed theory of language description,  Functional Generative Description \cite{meaningSgall1986} and was reflected in several annotation manuals available from the project website.\footnote{\url{https://ufal.mff.cuni.cz/pdt-c}} 

The multi-layer architecture (linked from meaning to text) allows a comprehensive description of the relations between morphological properties, syntactic function and expressed meaning, and thus contributes to greater accuracy in the description of the language and to the overall consistency of the annotated data (cf. \citealp{hajicova-etal-2022-advantages, mikulova-etal-2025-coling}). The multi-layer architecture is schematically illustrated in Fig.~\ref{fig:layers}: each annotation layer is indicated by a separate box. The links between the layers are indicated by the light dotted arrows. There are the three layers of annotation:

\begin{itemize}
\item \textbf{morphology} annotation (\textit{m-layer} box in Fig.~\ref{fig:layers}): all tokens  get a lemma and morphological tag (see Sect.~\ref{mlayer}).

\item \textbf{surface syntax} (\textit{a-layer}): a dependency tree capturing syntactic relations such as subject, object, adverbial, etc. (see Sect.~\ref{alayer}),

\item \textbf{deep syntax} and other \textbf{semantics} annotations (\textit{t-layer}), capturing  deep syntactic structure (Sect.~\ref{tlayer}), valency (\ref{valency}), coreference (Sect.~\ref{coreference}), discourse (Sect.~\ref{discourse}), etc.
\end{itemize}

In addition to the above-mentioned three annotation layers in the PDT scheme, there is also the \textbf{raw text layer} (it is not shown in Fig.~\ref{fig:layers}), where the text is segmented into documents and paragraphs and individual tokens are assigned unique identifiers. There is additional audio signal and speech recognition layer in the spoken data (Sect.~\ref{spoken}). In the spoken data part, the raw text layer is in fact also an ``annotated'' layer, namely the manually provided transcription of the audio signal. 

\textbf{Linking the layers}. To avoid losing any of the original information, tokens (nodes) at a lower layer are explicitly referenced from the closest (immediately higher) layer. These links enable every unit of annotation to be traced all the way down to the original text or transcript and audio (in spoken data).

\begin{table*}[t!]
\centering
\begin{tabular}{l|rrrrr}
                     & Written   & Translated & Spoken   & User-generated & Total
\\\hline\hline
Morphological layer  & 1,957,150 &  1,152,289 &  742,316 & 33,836  & 3,885,591
\\
Surface syntactic layer & 1,503,637 &  1,152,289 &  742,316 & 33,836 & 3,432,078
\\
Deep syntactic layer    & 833,180  &  1,152,289 &  742,316 & 33,836 & 2,761,621
\\
\bottomrule
\end{tabular}
\caption{Volume of the datasets in PDT-C 2.0 (number of tokens)}
\label{tab:volume}
\end{table*}

\section{Genre Diversified Data}
\label{data}

PDT-C 2.0 consists of four different datasets: %
written texts (Sect.~\ref{written}), translated texts (Sect.~\ref{translated}), spoken texts (Sect.~\ref{spoken}), and of user-generated texts (Sect.~\ref{faust}). 

\looseness-1
The datasets are uniformly published in three \textbf{formats}: pml, mrp, and treex. The Prague Markup Language format (PML, \citealp{pml2008}) is a language-independent, XML-based format customized for multi-layer linguistic annotation. Treex is technically also a PML format, used in the NLP system Treex (all annotation layers are in a single file; \citealp{ZabokrtskTreex}). MRP is a JSON-based format used in the CoNLL 2019 and 2020 shared tasks on meaning representation parsing \cite{oepen-etal-2019-mrp,oepen-etal-2020-mrp}; unlike the PML and Treex formats, the conversion to the MRP format, described in detail in \citet{zeman-hajic-2020-fgd}, is lossy because it extracts only part of the annotation.

\textbf{Quality and consistency} of the annotations were monitored, measured, and ensured using various tools (such as multiple annotations and automated checks; cf. \citealp{mikulova-stepanek-2010-ways,mikulova-etal-2022-quality,mikulova-etal-2025-coling}).

\subsection{Written Data}
\label{written}
The dataset of written texts coming from the \textbf{Prague Dependency Treebank}, the first PDT corpus, in development since the 1990s \cite{Hajic1998}. The data consist of Czech newspaper and journal texts from three domains: daily news, business, and science. Compared to other datasets, the annotation in the written dataset is the richest one, some special annotations are added; see Tab.~\ref{tab:annot}.

\subsection{Translated Data}
\label{translated}
The dataset of translated texts comes from the \textbf{Prague Czech-English Dependency Treebank} (PCEDT, originally published in 2012, \citealp{announcing2012}). PCEDT is a (partially) manually annotated Czech-English parallel corpus. The English part consists of the Wall Street Journal sections of the Penn Treebank \cite{marcus-etal-1993-building}. %
The Czech part, used in the PDT-C consolidated edition, has been manually (and professionally, with multiple quality control passes) translated from the English original, sentence to sentence.

\subsection{Spoken Data}
\label{spoken}
The dataset of spoken texts is taken from the \textbf{Prague Dependency Treebank of Spoken Czech} (originally published in 2017; \citealp{pdtsc20-2017}). It contains slightly moderated testimonies of Holocaust survivors from the Shoa Foundation Visual History Archive
and dialogues in which two participants chat over a collection of photographs.%

The spoken data differs from the other included PDT-corpora mainly in the “spoken” part of the corpus. In addition to the three annotation layers described in Sect.~\ref{layers}, the corpus also contains audio signal, transcript produced by an automatic speech recognition engine, and manual transcription of the recorded speech. The process starts at the “audio” layer, which contains the audio signal. The next layer contains the transcript as produced by an automatic speech recognition engine. The word layer contains manual transcription of the recorded speech, and the morphological layer contains the reconstructed, i.e. grammatically corrected version of the sentences (see Sect.~\ref{speech}). From this point on, annotation on the upper layers is standard. 

\subsection{User-generated Data}
\label{faust}
The dataset of user-generated texts comes from the \textbf{PDT-Faust} corpus, which is a small treebank containing short segments (very often with non-standard as well as expressive, obscene, or vulgar content) typed in by various users on the \url{reverso.net} web page for translation. The Czech data includes manual annotations of Czech reference translations of English source texts. This texts were translated independently by three translators and all three reference translations were annotated.

\section{Volume of Data}
\label{volume}

The data volume is given in Tab.~\ref{tab:volume}. Altogether, the consolidated treebank contains of almost 4 million tokens with manual morphological annotation (Sect.~\ref{mlayer}) and 3.5 million tokens with manual surface syntactic annotation (Sect.~\ref{alayer}) and 2.7 million with manual deep syntactic and  other semantic annotations (Sect.~\ref{tlayer}). The different number of tokens in the case of the written data is due to the fact that some annotations are only available for the morphological and/or surface syntactic layer.

\begin{table*}[t]
\begin{center}
\begin{tabular}{lllll}
Dataset/Type of annotation & Written & Translated & Spoken & User-generated
\\\hline\hline
Audio  & non-applicable & non-applicable & provided &
non-applicable
\\
ASR transcript & non-applicable & non-applicable & provided &
non-applicable
\\
Transcript & non-applicable & non-applicable & manually &
non-applicable
\\
Translation & non-applicable & manually & non-applicable &
manually
\\
\multicolumn{5}{c}{Morphological layer\vbox{\vskip 1.5em}}
\\\hline
Speech reconstruction & non-applicable & non-applicable & manually &
non-applicable
\\
Lemmatization & manually & manually & manually & manually
\\
Tagging & manually & manually & manually & manually
\\
\multicolumn{5}{c}{Surface syntactic layer\vbox{\vskip 1.5em}}
\\\hline
\textbf{Dependency structure} & manually & \textbf{manually} & \textbf{manually} &
\textbf{manually}
\\
\textbf{Surface syntactic functions} & manually & \textbf{manually} & \textbf{manually} &
\textbf{manually}
\\
Clause segmentation & manually & not annotated & not annotated & not
annotated
\\
\multicolumn{5}{c}{Deep syntactic layer\vbox{\vskip 1.5em}}
\\\hline
Deep syntactic structure & manually & manually & manually & manually
\\
Deep syntactic functions & manually & manually & manually & manually
\\
Valency  & manually & manually & manually & manually
\\
\textbf{Coreference} & manually & manually & manually & \textbf{manually}
\\
\textbf{Discourse} & manually & \textbf{manually} & \textbf{manually} & \textbf{manually}
\\
Grammatemes & manually & not annotated & not annotated & not annotated
\\
Topic-focus articulation & manually & not annotated & not annotated &
not annotated
\\
Bridging relations & manually & not annotated & not annotated & not
annotated
\\

Genre specification & manually & not annotated & not annotated & not
annotated
\\
Quotation & manually & not annotated & not annotated & not
annotated
\\
Multiword expressions & manually & not annotated & not annotated & not
annotated\\
\bottomrule
\end{tabular}
\caption{Overview of various types of annotation and their realization
  in the datasets (new manual annotation made to PDT-C 2.0 is indicated in bold)}
\label{tab:annot}
\end{center}
\end{table*}

\section{Rich Linguistic Annotation}
\label{rich}
The long-run Prague Dependency Treebank project is unique in its attempt to systematically cover and link different layers of language description including a rich semantic annotation.
Tab.~\ref{tab:annot} provides an overview of the different types of annotation across the three annotation layers (see Sect.~\ref{layers}) for each dataset (see Sect.~\ref{data}), along with information on how the annotations were carried out. Newly added manual annotations in the PDT-C 2.0 version are highlighted in bold. The table shows that all datasets include manual annotations for lemmatization, tagging, dependency structure, deep syntactic structure, valency, coreference, and discourse. In the spoken and written datasets, there are also additional specialized annotations. In the following subsections, the annotations of the most important phenomena are shortly described.

\subsection{Speech Reconstruction}
\label{speech}

Spontaneous speech reconstruction is a special type of manual annotation at the morphological layer that only belongs to the spoken data. The purpose of speech reconstruction is to ``translate'' the ``ungrammatical'' spontaneous speech to a written text, before it is tagged and parsed. The transcript is divided into sentence-like segments and the segments are edited to meet written-text standards, which means cleansing the text from the discourse-irrelevant and content-less material (e.g., superfluous words, false starts, repetitions, etc. are removed) and re-building the original segments into grammatical sentences with acceptable word order and proper morphosyntactic relations between words. %
See more in  \citet{PDTSCgoa2008} and \citet{pdtsc20-2017}.

\subsection{Lemmatization and Tagging}
\label{mlayer}

At the morphological layer, a lemma and a tag is assigned to each token. Czech is a highly inflectional language. A 15-character tag describes the inflectional forms of (declined) nouns and adjectives and (conjugated) verbs. All tokens of a sentence are traditionally also assigned a POS category within the tag. The annotation contains no syntactic structure, no attempt is made to put together analytical verb forms or other types of multiword expressions. The annotation is described in  \citet{novymanual,pdtc10}.

\begin{table}[h]
\small
\begin{center}
\begin{tabular}{lr}
COREFERENCE & count\\
\hline\hline
grammatical coreference & 66,168\\
textual coreference & 331,783\\
segment of text & 5,903\\
situational context & 20,753\\
\bottomrule
\end{tabular}
\caption{Volume of coreference annotations}
\label{tab:coreference}
\end{center}
\end{table}

\subsection{Surface Syntactic Annotation}
\label{alayer}

A surface dependency structure is captured by a rooted tree with the specification of the head for each node and the assignment of a syntactic function such as subject ({\tt Sb}), object ({\tt Obj}),  adverbial ({\tt Adv}), or attribute ({\tt Atr}). Every token of the raw text (including punctuation marks; cf. nodes for comma ({\tt AuxX}) and terminal symbol of the sentence ({\tt AuxK}) in Fig.~\ref{fig:layers}) is represented by a node and no additional nodes are allowed. The annotation guidelines are described in \citet{2026AManual}.
 
\subsection{Deep Syntactic Structure}
\label{tlayer}

At the deep syntactic layer, every sentence is represented as a tree-like graph. Unlike the lower layers, not all of the original tokens are represented as nodes, but the nodes only stand for content words (e.g., there is only one node for the preposition phrase \textit{s tím} ‘with that’ at the t-layer in Fig.~\ref{fig:layers}). Function words (prepositions, auxiliaries, etc.) do not have nodes of their own, their contribution to the meaning of the sentence is captured by several attributes attached to the nodes, the values of which represent this contribution (e.g., tense for verbs; see Sect.~\ref{gammatemes}). In case of surface deletions, extra nodes are added (in Fig.~\ref{fig:layers}, the restoration of a deletion is illustrated by the \texttt{\#PersPron} (personal pronoun) node for the Actor ({\tt ACT}) of the second sentence's predicate). The types of the (semantic) dependency relations are represented by the \textit{functor} attribute attached to all nodes. Annotation principles are described in several manuals \cite{manualtr2006,Annotationtectogrammatical2014}.

\begin{table}[ht!]
\small
\begin{center}
\begin{tabular}{lr}
DISCOURSE & Count\\
\midrule\midrule
Discourse relations: COMPARISON\\
\midrule
concession & 2,736\\
confrontation & 2,432\\
correction & 1,390\\
gradation & 1,108\\
opposition & 10,348\\
pragmatic contrast & 231\\
restrictive opposition & 1,019\\
\midrule
Discourse relations: CONTINGENCY\\
\midrule
condition & 4,801\\
explication & 373\\
pragmatic condition & 372\\
pragmatic reason-result & 493\\
purpose & 2,710\\
reason--result & 10,927\\
\midrule
Discourse relations: EXPANSION\\
\midrule
conjunction & 29,509\\
conjunctive alternative & 854\\
disjunctive alternative & 623\\
equivalence & 544\\
generalization & 522\\
instantiation & 687\\
specification & 1,428\\
\midrule
Discourse relations: TEMPORAL\\
\midrule
precedence--succession & 6,600\\
synchrony & 2,799\\
\midrule
DISCOURSE RELATIONS: TOTAL & 82,506\\
\bottomrule
\end{tabular}
\caption{Volume of discourse annotations}
\label{tab:discource}
\end{center}
\end{table}

\subsection{Valency}
\label{valency}

The core ingredient in the annotation of deep structure is valency (predicate-argument structure annotation). The valency criterion divides functors into argument and adjunct functors. There are five arguments: Actor ({\tt ACT}), Patient ({\tt PAT}), Addressee ({\tt ADDR}), Origin ({\tt ORIG}), and Effect ({\tt EFF}). In addition, about 50 types of adjuncts (temporal, spatial, manner, causal, regard, etc.) are used. For a particular verb (or more precisely, verb sense), a subset of the functors is obligatory, while others are either not present at all or are optional. Each occurrence of a verb in all corpora is linked to the appropriate valency frame in the valency lexicon (see Sect.~\ref{pdtvallex}). 

\subsection{Coreference}
\label{coreference}

\looseness-1
Coreference annotation \cite{hajicova-etal-2000-coreference} captures a referential identity relation between entities (nodes; ex. \textit{Einstein} - \textit{he} - \textit{the famous scientist}). Several types are distinguished. Grammatical pronominal coreference is based on language-specific grammatical rules, whereas resolving textual coreference (both pronominal and nominal) requires contextual knowledge. Textual coreference annotation follows the "chain principle", where the anaphoric entity always refers to the last preceding antecedent. 
Coreference can also be cataphoric (pointing to a subsequent part of the text). Two special cases of coreference are further annotated: reference to situational context and reference to a segment of text. In Fig.~\ref{fig:layers}, coreference relations are represented by the brown (grammatical) and blue (textual) arrows. In PDT-C 2.0, there is now manual coreference annotation in all four datasets; the volume of the coreference annotations is in Tab.~\ref{tab:coreference}.

\subsection{Discourse}
\label{discourse}

Annotation of discourse relations covers local relations that hold between two spans of text (usually clauses and sentences) marked by primary or secondary discourse connectives.
Primary connectives are grammaticalized, mostly one-word expressions (such as \emph{a} `and', \emph{ale} `but', \emph{protože} `because'), secondary connectives are more loose expressions (\emph{z toho důvodu} `for that reason', \emph{na druhou stranu} `on the other hand', etc.). Each relation is accompanied by a discourse type (such as {\tt reason--result}, {\tt condition}, {\tt purpose}, {\tt equivalence}, etc.; see Tab.~\ref{tab:discource} for a complete list). See more in \citet{discourse-manual2012,mirovsky-etal-2024-cost}. Discourse relations are annotated between roots of the relevant subtrees (sentences, clauses, phrases) in the trees. In Fig.~\ref{fig:layers}, a discourse relation (of {\tt confrontation}) is represented by the orange arrow between the predicates of the sentences. In PDT-C 2.0, there is now manual discourse annotation in all four datasets.

\subsection{Topic-Focus Articulation}
A basic aspect of the deep structure is also the topic-focus articulation (for arguments on its semantic relevance see \citealp{meaningSgall1986,topicfocusarticulation1998}), %
indicated by the blue values {\tt t} and {\tt f}  (in front of the functor values) in Fig.~\ref{fig:layers}: {\tt t} is for contextually bound and {\tt f} for contextually non-bound nodes. The ordering of nodes corresponds to the information structure of a sentence (cf. different position of particle \textit{stěží} ‘hardly’ and \textit{ale} ‘but’ at the a-layer and t-layer in Fig.~\ref{fig:layers}).\footnote{The nodes at the lower layers are naturally ordered based on the surface word order.} In PDT-C 2.0, topic-focus articulation is captured only in the written dataset (cf. Tab.~\ref{tab:annot}).   

\subsection{Grammatemes}
\label{gammatemes}
So called grammatemes \cite{grammatemes2006,morphological2010}) are attached to some nodes; they provide information about the node that cannot be derived from the deep syntactic structure, the functor and other attributes. Grammatemes are counterparts of those morphological categories which bear relevant semantic information (e.g., tense of predicate, number of entities, modality of the sentence). They are annotated only in the written dataset (cf. Tab.~\ref{tab:annot}).

\subsection{Bridging}
\label{bridging}
In the written dataset, apart from the coreference relations, non-coreferential association relations are annotated as bridging relations if they are related in one of specific types of semantic or conceptual ways to their antecedents. Several types are distinguished, e.g., metonymical relation between a part and a whole ({\tt part-of}; ex. \textit{room – ceiling}); relation between a set and its subsets ({\tt set-subset}, ex. \textit{students – some students – a student}), the relation between an entity and a singular function on this entity ({\tt function}; ex. \textit{prime minister – government}). 
See more in \citet{extended2011}.

\subsection{Other Annotation}
\label{other}
In the written dataset, noun valency, and other semantic properties of the sentence such as genre specification, multiword expressions, quotation are also annotated. More information of these special annotations can be found in \citet{fromPDT2013}.

\section{External Resources}
\label{resources}

An important part of annotation also involves various dictionaries. They can be used to distinguish the different meanings of words and also to maintain or monitor the consistency of annotations. The PDT-C annotation is associated with the morphological (Sect.~\ref{morfflex}) and valency (\ref{pdtvallex}) dictionaries.

\subsection{MorfFlex}
\label{morfflex} 

MorfFlex (the latest version is \textbf{MorfFlex CZ 2.1}, \citealplanguageresource{morfflex}, \citealp{morfflex-2026}) is the Czech morphological dictionary. For each word form, full inflectional information is coded in a positional tag. Word forms are organized into paradigms according to their morphological behaviour. The paradigm is identified by a unique lemma. The description also contains some semantic, stylistic and derivational information. MorfFlex is distributed as a flat list of \textit{form - lemma - tag} triplets. MorfFlex CZ 2.1 contains 126,906,921 such triplets. It is fully compatible with the PDT-C 2.0 morphological annotation.

\begin{table*}[t]
  \centering
    \small
  \setlength{\tabcolsep}{2.8pt}
  \catcode`! = 13\def!{\itshape}
  \begin{tabular}{lcccccccccccc}
    \toprule
      Tags & \multicolumn{5}{c}{Tagging Accuracy}
        & \multicolumn{5}{c}{Lemmatization Accuracy} & \multicolumn{2}{c}{Performance} \\\cmidrule(lr){2-6}\cmidrule(lr){7-11}\cmidrule(lr){12-13}
      Predicted & Faust & PCEDT & PDT & PDTSC &\kern-.3em MacroAvg & Faust & PCEDT & PDT & PDTSC &\kern-.3em MacroAvg
        & Speed & Size \\
    \midrule
      Full tags   & 95.08 & 97.24 & 95.95 & 97.63 & 96.47 & 97.69 & 98.98 & 98.29 & 98.83 & 98.45 & 26k$\scriptsize\frac{\textrm{toks}}{\textrm{s}}$ & 25MB \\
     !First 2 pos &!98.24 &!99.30 &!98.61 &!99.10 &!98.81 &!97.31 &!98.79 &!98.10 &!98.69 &!98.22 &!\llap{3}28k$\scriptsize\frac{\textrm{toks}}{\textrm{s}}$ &!10MB \\
    \bottomrule
  \end{tabular}
  \caption{Performance of morphological tagging and lemmatization of two MorphoDiTa tagger models,
  one predicting full tags (15 positions) and the other predicting only first
  2 tag positions.}
  \label{tab:morphodita_results}
\end{table*}

\begin{table*}[t]
  \vspace{5pt}
  \centering
    \small
  \setlength{\tabcolsep}{3.8pt}
  \begin{tabular}{lccccccccccc}
    \toprule
      \multirow{2}[2]{*}{Test Data} & \multicolumn{2}{c}{Morpho Train}  & \multicolumn{2}{c}{Syntax Train}
        & \multicolumn{2}{c}{Raw Predictions} & \multicolumn{3}{c}{With MorphoDiTa}
        & \multirow{2}[2]{*}{UAS}
        & \multirow{2}[2]{*}{LAS} \\\cmidrule(lr){2-3}\cmidrule(lr){4-5}\cmidrule(lr){6-7}\cmidrule(lr){8-10}
      & PDT & PDT-C & PDT & PDT-C & Tags & Lemmas & Tags & Lemmas &\kern-.2em LemmasEM\kern-.2em \\
    \midrule\multirow{3}{*}{PDT-C: Faust}
      & \YES & \NO & \YES & \NO & 95.64 & 96.36 & 97.02 & 98.22 & 97.85 & 83.72 & 78.94 \\
      & \NO & \YES & \YES & \NO & 97.13 & 98.27 & 97.77 & 98.91 & 98.46 & 84.20 & 79.97 \\
      & \NO & \YES & \NO & \YES & 97.18 & 98.24 & 97.82 & 99.15 & 98.70 & 91.44 & 88.40 \\
    \noalign{\kern4pt}\multirow{3}{*}{PDT-C: PCEDT}
      & \YES & \NO & \YES & \NO & 98.11 & 98.46 & 98.44 & 98.94 & 98.40 & 95.57 & 93.63 \\
      & \NO & \YES & \YES & \NO & 98.84 & 99.25 & 99.00 & 99.51 & 98.86 & 95.74 & 93.86 \\
      & \NO & \YES & \NO & \YES & 98.89 & 99.27 & 99.06 & 99.55 & 98.90 & 96.52 & 94.97 \\
    \noalign{\kern4pt}\multirow{3}{*}{PDT-C: PDT}
      & \YES & \NO & \YES & \NO & 98.18 & 98.39 & 98.49 & 98.92 & 98.73 & 95.17 & 92.63 \\
      & \NO & \YES & \YES & \NO & 98.26 & 98.50 & 98.49 & 98.93 & 98.73 & 95.25 & 92.76 \\
      & \NO & \YES & \NO & \YES & 98.20 & 98.39 & 98.45 & 98.86 & 98.67 & 95.24 & 92.80 \\
    \noalign{\kern4pt}\multirow{3}{*}{PDT-C: PDTSC}
      & \YES & \NO & \YES & \NO & 97.30 & 97.67 & 97.86 & 98.42 & 98.36 & 93.60 & 89.97 \\
      & \NO & \YES & \YES & \NO & 98.68 & 99.08 & 98.91 & 99.34 & 99.27 & 93.69 & 90.08 \\
      & \NO & \YES & \NO & \YES & 98.71 & 99.08 & 98.96 & 99.35 & 99.29 & 94.96 & 92.33 \\
    \midrule\multirow{3}{*}{PDT-C: MacroAvg}
      & \YES & \NO & \YES & \NO & 97.31 & 97.72 & 97.95 & 98.62 & 98.34 & 92.01 & 88.79 \\
      & \NO & \YES & \YES & \NO & 98.23 & 98.77 & 98.54 & 99.17 & 98.83 & 92.22 & 89.17 \\
      & \NO & \YES & \NO & \YES & 98.24 & 98.74 & 98.57 & 99.23 & 98.89 & 94.54 & 92.12 \\
    \bottomrule
  \end{tabular}
  \caption{Comparison of UDPipe morphosyntactic performance in percents using
  either whole PDT-C or just PDT as morphological/syntactic annotation training data,
  evaluated on its four subsets, with or without MorphoDiTa dictionary during
  inference. Predicted lemma is considered correct when a raw lemma plus an
  optional lemma sense match; LemmasEM compares full lemmas including all
  additional information.}
  \label{tab:udpipe_results}
\end{table*}

\subsection{PDT-Vallex}
\label{pdtvallex}

The valency dictionary PDT-Vallex  \cite{PDTVALLEXCreating2003,BuildingPDTVALLEX2012} was developed in parallel with the annotation and contains almost exclusively verbs and their meanings that occurred in the annotated data, whose valency the annotators needed to know in order to correctly annotate valency. The latest version, \textbf{PDT-Vallex 4.5} \citelanguageresource{pdtvallex}, includes nearly 8,500 verbal lemmas and 14,500 valency frames. It is part of PDT-C 2.0 release, and the valency frames are directly linked to the data through explicit references.

\section{Related Data and Tools}
\label{related}

Throughout their development, the PDT corpora have served as an invaluable resource for linguistic research, for enriching the description of the Czech language system, and for developing general methods of language description.
The richly linguistically annotated PDT corpora are also widely used in international comparisons in the NLP field. In what follows, we briefly introduce the latest morphological, syntactic, and semantic analysers (Sect.~\ref{tools}), as well as the most recent conversions of PDT-C into other formalisms (Sect.~\ref{conversions}).

\subsection{Tools}
\label{tools}

We now describe and evaluate morphological, morphosyntactic and deep syntactic
analyzers trained on PDT-C 2.0; where appropriate, we compare them to models
trained only on PDT syntactic data, the only available syntactic data
in PDT-C 1.0. All models are released under the CC BY-NC-SA license.

\paragraph{MorphoDiTa} We train a fast CPU-based tagger and lemmatizer using
MorphoDiTa~\citep{strakova14}. The performance of the resulting model is presented in
Table~\ref{tab:morphodita_results} and the model is available at
{\small\url{https://hdl.handle.net/11234/1-5985}}.

\begin{table*}[t]
  \centering
    \small
  \setlength{\tabcolsep}{3.6pt}
  \begin{tabular}{lccccccccccc}
    \toprule
      \multirow{2}[2]{*}{Test Data} & \multicolumn{2}{c}{Syntax Train}  & \multicolumn{2}{c}{MRP Train}
        & \multirow{2}[2]{*}{Labels}
        & \multirow{2}[2]{*}{Properties}
        & \multirow{2}[2]{*}{Anchors}
        & \multirow{2}[2]{*}{Edges}
        & \multirow{2}[2]{*}{Attributes}
        & \multirow{2}[2]{*}{Tops}
        & \multirow{2}[2]{*}{\textbf{All}} \\\cmidrule(lr){2-3}\cmidrule(lr){4-5}
      & PDT & PDT-C & PDT & PDT-C \\
    \midrule\multirow{3}{*}{PDT-C: Faust}
      & \YES & \NO & \YES & \NO & 88.74 & 51.72 & 90.77 & 56.30 & 54.28 & 92.00 & 66.28 \\
      & \NO & \YES & \YES & \NO & 92.30 & 85.32 & 94.04 & 77.79 & 75.24 & 99.00 & 85.68 \\
      & \NO & \YES & \NO & \YES & 92.48 & 85.88 & 95.01 & 77.67 & 76.25 & 96.67 & 86.01 \\
    \noalign{\kern4pt}\multirow{3}{*}{PDT-C: PCEDT}
      & \YES & \NO & \YES & \NO & 94.99 & 53.17 & 96.62 & 67.45 & 59.04 & 99.83 & 72.97 \\
      & \NO & \YES & \YES & \NO & 96.97 & 92.95 & 97.91 & 87.19 & 80.27 & \llap{1}00.00 & 92.79 \\
      & \NO & \YES & \NO & \YES & 96.96 & 92.18 & 98.02 & 87.39 & 81.30 & 99.83 & 92.71 \\
    \noalign{\kern4pt}\multirow{3}{*}{PDT-C: PDT}
      & \YES & \NO & \YES & \NO & 96.55 & 92.87 & 97.66 & 82.41 & 74.76 & 99.09 & 91.36 \\
      & \NO & \YES & \YES & \NO & 96.62 & 88.78 & 97.79 & 81.87 & 75.57 & 99.93 & 89.56 \\
      & \NO & \YES & \NO & \YES & 96.39 & 89.20 & 97.85 & 82.68 & 77.42 & 99.56 & 89.93 \\
    \noalign{\kern4pt}\multirow{3}{*}{PDT-C: PDTSC}
      & \YES & \NO & \YES & \NO & 93.21 & 52.72 & 97.05 & 67.50 & 72.82 & 99.60 & 72.48 \\
      & \NO & \YES & \YES & \NO & 96.22 & 92.64 & 98.30 & 86.87 & 83.87 & 99.95 & 92.59 \\
      & \NO & \YES & \NO & \YES & 96.24 & 92.78 & 98.30 & 87.54 & 84.72 & 99.60 & 92.85 \\
    \midrule\multirow{3}{*}{PDT-C: MacroAvg}
      & \YES & \NO & \YES & \NO & 93.37 & 62.62 & 95.52 & 68.41 & 65.22 & 97.63 & 75.77 \\
      & \NO & \YES & \YES & \NO & 95.53 & 89.92 & 97.01 & 83.43 & 78.74 & 99.72 & 90.16 \\
      & \NO & \YES & \NO & \YES & 95.52 & 90.01 & 97.30 & 83.82 & 79.92 & 98.91 & 90.38 \\
    \bottomrule
  \end{tabular}
  \caption{Comparison of PERIN semantic parsing MRP score in percents on the
  four PDT-C subsets, evaluated using either whole PDT-C or just PDT as
  syntactic/semantic annotation training data.}
  \label{tab:perin_results}
\end{table*}

\paragraph{UDPipe} We also train morphosyntactic parser model using
a GPU-based UDPipe~\citep{straka-2018-udpipe,straka-etal-2019-udpipe}. Notably,
we utilize the variant described by~\citet{straka-strakova-open-source-2024}
which improves performance by consulting the morphological dictionary provided
by MorphoDiTa during inference.

Table~\ref{tab:udpipe_results} shows the performance of three UDPipe models:
trained only on the PDT data, trained on full PDT-C morphological data but
only on PDT syntactic data (best configuration trainable on PDT-C
1.0;~\citealp{straka-strakova-open-source-2024}), and trained on the full PDT-C
2.0 data. Using the latter model reduces the macro-averaged error rate in
syntactic parsing accuracy by more than 27\%.

Compared to the MorphoDiTa tagger, UDPipe achieves 60\% error reduction in
tagging accuracy and 50\% in lemmatization; however, MorphoDiTa model is by
a decimal order of magnitude faster on a single CPU.

The model trained on full PDT-C 2.0 data is available at
{\small\url{http://hdl.handle.net/11234/1-6045}}.

\paragraph{PERIN} Finally, we train a meaning representation parser producing
semantic graphs~\citep{zeman-hajic-2020-fgd} using
PERIN~\citep{samuel-straka-2020-ufal}, the winning system of the 2020 CoNLL
shared task on Cross-Framework Meaning Representation Parsing (MRP 2020;
\citealp{oepen-etal-2020-mrp}).

To produce a semantic graph, the PERIN parser processes not just an input
sentence but also its syntactic tree; therefore, it builds on top of a UDPipe
parser. We therefore train three parser configurations: first training both
UDPipe and PERIN on PDT data (the setting of MRP 2020), then training UDPipe on
PDT data and PERIN on PDT-C data (the best setting in PDT-C 1.0), and finally
using the complete PDT-C data for both systems.

The results of all configurations is presented in
Table~\ref{tab:perin_results}. When using just PDT to train the semantic
parser, the results on the other subsets are considerably worse; on the other
hand, using all PDT-C training data decreases performance on PDT slightly
(indicating either limited parser capacity or mild annotation differences).

Maybe surprisingly, when using all PDT-C data to train the semantic parser,
the quality of the syntactic trees has only minor influence on the final results,
i.e., even with slightly lower-quality trees on non-PDT subsets, PERIN
successfully learns to produce high-quality semantic graphs from them.

The model trained on full PDT-C 2.0 data is available at
{\small\url{http://hdl.handle.net/11234/1-6118}}.

\subsection{Conversion to Other Frameworks}
\label{conversions}

An important application (and promotion) of an annotated corpus is its conversion into another formalisms. By being incorporated into different frameworks, the robustness and universality of the chosen format become evident. The PDT corpora are used for conversions into various frameworks. 

First, it is important to mention the popular \textbf{Universal Dependencies (UD)} formalism \cite{de-marneffe-etal-2021-universal}. After the conversion of PDT-C 2.0 (in UD version 2.16),\footnote{\url{http://hdl.handle.net/11234/1-5901}} Czech represents the largest language coverage \cite{ud-pdtc-2026}.

The PDT corpora are part of other expanding UD projects: \textbf{DeepUD}\footnote{\url{http://hdl.handle.net/11234/1-3720}} \cite{deepUD2019} enriches the basic UD annotation with deep syntactic annotations; the \textbf{CorefUD} project\footnote{\url{ http://hdl.handle.net/11234/1-5896}} \cite{corefUD-2022} is an initiative for harmonizing coreference corpora into a unified format. 

\looseness1
The discourse annotation in the PDT-C 2.0 is also converted into the \textbf{Penn Discourse Treebank  format}
\cite{prasad-etal-2008-penn,Mirovsky2023} within the \textbf{Prague Discourse Treebank 4.0} (PDiT 4.0)\footnote{\url{http://hdl.handle.net/11234/1-5680}} release \cite{MiSyPresentingPDiT40-2026}.

The rich annotation at the t-layer serves as a source for conversion into various semantic and knowledge representations. Among earlier conversions, let us mention the transformation into the formal-logical format \textbf{Minimal Recursion Semantics} (\citealp{copestake2005minimal,jakob-etal-2010-mapping}). Most recently, the PDT-C 2.0 data have been converted into the \textbf{Uniform Meaning Representation format} \cite{umr2021,itatumr2024}.\footnote{\url{http://hdl.handle.net/11234/1-5951}}

\section{Future Work}
\label{future}

The description of language is far from complete. Despite the remarkable success of large language models (LLMs), we are still far from achieving systems that truly understand natural language, and fundamental linguistic research remains essential. In this respect, we aim to continue our efforts in systematically describing language from form to meaning. For the next version, PDT-C 3.0, the following annotation efforts have already been initiated: (i) event-type annotation \cite{Linking2025,synsemclassLREC2026}, (ii) implicit discourse relations \cite{implicit2019}, and (iii) fine-grained classification of (circumstantial) semantic roles, such as temporal, spatial, manner, and causal roles \cite{mikulova-2024-fine}.
These extensions will further enrich the corpus, enabling deeper linguistic analysis and providing a robust foundation for future NLP research and applications.

\section{Conclusion}
\label{conclusion}

In the contribution, we present the Prague Dependency Treebank – Consolidated 2.0, a comprehensive, multi-layer linguistic resource that integrates semantic, syntactic, and morphological information, including inter-sentential phenomena such as coreference and discourse relations. The long-term development of the PDT framework has resulted in a uniformly annotated, genre-diversified corpus of almost 4 million Czech tokens, supported by fully compatible lexicons (morphological and valency). Its rich annotation makes it an invaluable tool for both linguistic research and the development of traditional and advanced NLP applications, as well as for conversion into many other formats across all levels of linguistic description. The corpus is freely available for use.

\section{Limitations}
While we present a large, genre-diversified, and richly annotated language resource, there are several limitations we are aware of. Some types of annotation are currently available only for a subset of the treebank — this applies in particular to grammatemes (Sect.~\ref{gammatemes}), as well as to bridging relations (Sect.~\ref{bridging}) and multiword expressions (Sect.~\ref{other}); cf. Tab.~\ref{tab:annot}. Moreover, the treebank is monolingual. Although an English counterpart exists, it is released separately as part of a parallel dataset (cf. Sect.~\ref{translated}). The new release of the Czech-English parallel treebank (PCEDT 3.0) is planned for late 2026.

The evaluation of predicted semantic graphs is currently based on MRP
score~\citep{oepen-etal-2020-mrp}, which considers only the subset of annotation
converted to the MRP format~\citep{zeman-hajic-2020-fgd}. Evaluating the full
annotations will require to devise a new evaluation metric.

\section{Acknowledgements}
The research reported here has been supported by the Czech Science Foundation under the project 22-03269S. The work described herein has also been supported by the Ministry of Education, Youth and Sports of the Czech Republic, Project No. LM2023062 LINDAT/CLARIAH-CZ.\footnote{\url{https://lindat.cz}}

\section{Bibliographical References}\label{sec:reference}

\bibliographystyle{lrec2026-natbib}
\bibliography{lrec2026-example}

\section{Language Resource References}
\label{lr:ref}
\bibliographystylelanguageresource{lrec2026-natbib}
\bibliographylanguageresource{languageresource}

\end{document}